\documentclass[a4paper,conference]{IEEEtran}

\ifCLASSINFOpdf
\else
\fi
\DeclareUnicodeCharacter{2061}{}
\usepackage{graphicx}
\usepackage{cite}
\usepackage{amsmath,amssymb,amsfonts}
\usepackage{algorithmic}
\usepackage{textcomp}
\usepackage{xcolor}
\usepackage{graphicx}
\usepackage{caption}
\usepackage{float}
\usepackage{subcaption}
\usepackage{booktabs}
\usepackage{multirow}
\usepackage{comment}
\usepackage{lipsum}
\newcommand\blfootnote[1]{%
  \begingroup
  \renewcommand\thefootnote{}\footnote{#1}%
  \addtocounter{footnote}{-1}%
  \endgroup
}

\begin{document}
%
\title{ScreenSeg: On-Device Screenshot Layout Analysis}
\makeatletter
\newcommand{\linebreakand}{%
  \end{@IEEEauthorhalign}
  \hfill\mbox{}\par
  \mbox{}\hfill\begin{@IEEEauthorhalign}
}
\makeatother

\author{\IEEEauthorblockN{Manoj Goyal*}
\IEEEauthorblockA{OnDevice AI\\
Samsung R\& D Institute\\
Bangalore, India\\
manoj.goyal@samsung.com}
\and
\IEEEauthorblockN{Rachit S Munjal*}
\IEEEauthorblockA{OnDevice AI\\
Samsung R\& D Institute\\
Bangalore, India\\
rachit.m@samsung.com}
\and
\IEEEauthorblockN{Sukumar Moharana}
\IEEEauthorblockA{OnDevice AI\\
Samsung R\& D Institute\\
Bangalore, India\\
msukumar@samsung.com}
\linebreakand
\IEEEauthorblockN{Deepak Garg}
\IEEEauthorblockA{OnDevice AI\\
Samsung R\& D Institute\\
Bangalore, India\\
deepak.garg@samsung.com}
\and
\IEEEauthorblockN{Debi Prasanna Mohanty}
\IEEEauthorblockA{OnDevice AI\\
Samsung R\& D Institute\\
Bangalore, India\\
debi.m@samsung.com}
\and
\IEEEauthorblockN{Siva Prasad Thota}
\IEEEauthorblockA{OnDevice AI\\
Samsung R\& D Institute\\
Bangalore, India\\
siva.prasad@samsung.com}
}


\maketitle

\begin{abstract}
We\blfootnote{* Primary authors} propose a novel end-to-end solution that performs a Hierarchical Layout Analysis of screenshots and document images on resource constrained devices like mobilephones. Our approach segments entities like Grid, Image, Text and Icon blocks occurring in a screenshot. We provide an option for smart editing by auto highlighting these entities for saving or sharing. Further, this multi-level layout analysis of screenshots has many use cases including content extraction, keyword-based image search, style transfer, etc. We have addressed the limitations of known baseline approaches, supported a wide variety of semantically complex screenshots, and developed an approach that is highly optimized for on-device deployment. In addition, we present a novel weighted NMS technique for filtering object proposals. We achieve an average precision of about 0.95 with a latency of around 200ms on the Samsung Galaxy S10 Device for a screenshot of 1080p resolution. The solution pipeline is already commercialized in Samsung Device applications i.e. Samsung Capture, Smart Crop, My Filter in Camera Application, Bixby Touch.
\end{abstract}


%
\IEEEpeerreviewmaketitle

\section{Introduction}
Recent years have seen a significant increase in multimedia data in smartphones. A survey states that a user takes 9 screenshot per week on an average. So Screenshot Layout Analysis becomes an important problem to address.

Layout Analysis is the first major step to gain a complete semantic understanding of an image by comprehending the placement of constituent elements and building a knowledge representation for the image. Entity segmentation from an image is a well studied problem in computer vision and is a key pre-processing step in content extraction, retrieval, or mapping.

With the ever-increasing use of mobile devices, there is a growing need to analyze the contents on the screen and perform intelligent extraction of information. In this paper, we propose a novel mechanism to perform a Hierarchical Layout Analysis of Mobile Screenshots by combining Entity Segmentation and Image Processing techniques. The whole pipeline is kept light-weight to enable real-time, on-device layout understanding.

The paper is organized as follows: In Section 2, we discuss the related work done in the field of layout analysis. Section 3, presents our efficient layout analysis pipeline and describes each module in detail. In Section 4, we showcase the experimental results and performance of this pipeline. In Section 5, we give a logical conclusion to our work and propose some future work in this field.

\begin{figure*}[htbp]
\includegraphics[width=1\textwidth]{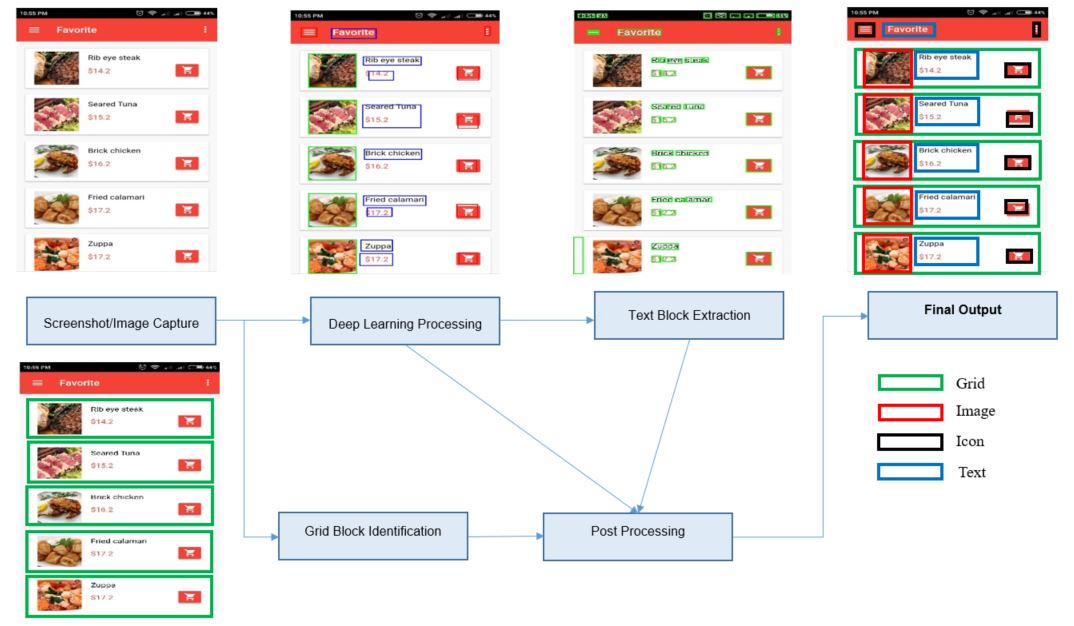}
\label{fig:pipeline}
\centering
\caption{Overview of Process Pipeline}
\centering
\end{figure*}

\section{Background}
 Layout Analysis is a well-known problem statement and has numerous applications, such as intelligent information retrieval, content understanding, content transformation, content tagging, etc. Despite its wide range of applications, the study of layout analysis has been limited to documents in the past. Many methods for understanding document images have been investigated to extract and classify meaningful information from documents. Classical methods of layout analysis [1] involve performing morphological operations, connected component analysis, and then classifying extracted features into their constituent regions. Recent advances in highly precise and robust deep neural networks have led to the exploration of deep learning-based layout analysis as well. DeepLayout [2], which uses a deep learning approach of semantic segmentation for layout understanding, outperforms some of the classical methods in the field. So far, the proposed methods for layout analysis are applicable to document images, which have a comparatively simple and limited set of the overall structure, constituent elements, and backgrounds.

  We extend the existing work and knowledge in this field to a new domain of mobile screenshots. Unlike documents, in case of layout analysis in mobile application screenshots, the overall structure and constituent elements are highly diverse and complex due to the ever-growing digital content and app layouts. Also, we have to deal with complex gradient backgrounds and develop a foreground-background understanding in such cases. Extraction of image block from the top of a background image or scene-text block from the top of an image block and on-device execution with constraint resources are some of the key challenges that we faced. We perform a multi-hierarchical layout analysis, by extraction of        1. Grid blocks, 2. Image, Text, and Icon blocks, 3. Scene Text blocks on top of image blocks.

We come up with a pipeline with a combination of Deep Neural Networks and Image Processing techniques to achieve this task with high accuracy and less inference time. The suggested approach is capable to handle all different type of screenshot images with complex backgrounds. The steps involved include 1) Deep learning based Image, Text, and Icon detection 2) Comparison of vision based and Deep Learning based approach for Text extraction 3) Grid Identification and Block Rectification. Each module is discussed in detail in the sections to follow.

To the best of our knowledge, by far no such methodology has been proposed to analyse layout of an application screenshot. We propose a unique, light-weight and highly optimized On-Device capable pipeline for layout analysis on application screenshots.

\section{Process Pipeline}
 
Fig. 1 shows the overview of the overall Process Pipeline. The pipeline, which contains a combination of deep learning and computer vision based modules, is constructed keeping in mind efficient resource usage, dependency on previous modules and need of real-time speed. The various components are discussed in detail below.
\subsection{Layout Analysis VIA DNN}
Images, Icons and Texts are the integral components and building blocks of any screenshots and detecting them is a major step in Screenshot Layout Analysis. We model this problem as an object detection task, in which the components of the screen can be treated as separate object classes. While Text and Icons have a lot of common features in between all the objects of their respective classes, Image class has a wide variety of sub-categories like Product Images, Natural Images, Posters, and Memes etc. This makes detecting all the different types of images, under a single category a challenging task. Gradient, Texture and Sharp color changes are some of the features that differentiate Image class from all the other classes.

A modified and reduced version of \textbf{SqueezeDet} [3] architecture is applied and trained on a hand-crafted dataset. Its result is profiled and evaluated against MobileNet SSD [4] architecture. Though both models showed almost similar accuracies in our case, SqueezeDet is preferred as a starting base network due to its low memory footprint and robust fire modules which led to lesser inference time. A resized and normalized input is fed to the network, which performs a single shot detection, first by extracting a high dimensional, low resolution feature map for the input image followed by ConvDet, a convolutional layer to take the feature map as input and compute bounding box proposals and their class probabilities. It works on a sliding window approach that moves through each spatial position on the feature map. A set of reference anchor boxes are defined of varying aspect ratio and sizes matching the data distribution, with predefined uniformly separated center points The output from the network is a vector of size CP * K × (4 + 1 + C) values that encode the bounding box predictions, confidence scores and class probabilities per center point per anchor box respectively. Here, CP denotes the number of center points, K denotes the number of anchor box shapes at each center point and C is the number of classes. The DNN details have been presented in Figure 2.

\begin{figure}[h!]
\includegraphics[width=1\linewidth]{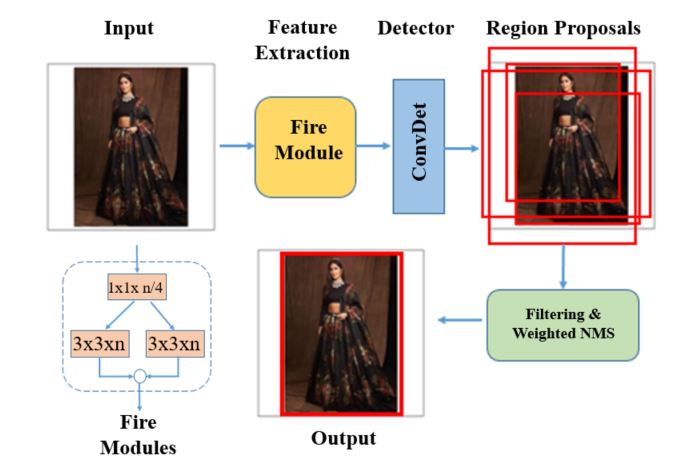}
\label{fig:LayoutDL}
\centering
\caption{DNN Architecure and Weighed NMS}
\centering
\end{figure}

We use a weighted multi-task optimization function, which comprises of bounding box regression loss, weighted cross entropy loss for classification and confidence score regression loss. For bounding box loss and confidence score loss, we use mean square error loss. The total loss is defined as:
\[L_{total}=L_{bbox}+{\alpha}.L_{class}+{\beta}.L_{conf}\]
\[L_{class} (y,y^p )=-\sum_{(j=0)}^M\sum_{(i=0)}^Ncw.y_{(i,j)}.log⁡(p_{(i,j)})\]
\[L_{bbox} (y,y^p) = {\frac{1}{B}} \sum_{(i=1)}^Bx_{(c,i)}^p-x_{(c,i)})^2+ (y_{(c,i)}^p-\]
\[  y_{(c,i)})^2+(w_i^p-w_i)^2+(h_i^p-h_i)^2\]
\[L_{conf} (y,y^p) = {\frac{1}{B}} \sum_{(i=1)}^B({\delta}_p-{\delta}_{GT})^2 \]

Where $\delta_p$ is the output from convDet layer and $\delta_{GT}$ is the IoU with the ground truth. $\alpha$ and $\beta$ are tuned empirically, cw are the class weights, B is the total number of anchor grids, $x_{c,i}$, $y_{c,i}$, $w_i$ and $h_i$  are the center points, width and height of the $i_{th}$ anchor grid.

\textbf{\emph{1) Datasets:}}
Any model can only be as good as its dataset. So, creating a high quality and diverse dataset is a crucial step in building any deep learning model. There are datasets available for layout analysis in documents and texts, but the dataset is not readily available for layout analysis of mobile screenshots. We created the dataset by capturing around 2000 unique mobile screenshots of different types of layouts. Further, each object instance i.e. Image, Text, Icon was annotated with its corresponding class and bounding box coordinates with the help of 3 trained annotators. We used data augmentation techniques by taking the annotated screenshots as templates and randomly superimposing images, text and icons to the respective annotated layout regions. Apart from this, synthetic layouts were also generated and used for augmentation to generate a total of 25,000 images for training and validation.

\textbf{\emph{2) Class Imbalance Handling:}}
Mobile screens are usually populated by texts and icons, i.e. on average the number of text and icon blocks per screenshot is 15-20 times more than the number of image blocks. This leads to a huge class imbalance [18] in the data. The icon and text class overwhelm the image class, which leads the model to give poor results for image class if this problem is not addressed properly.

  We handle the class imbalance problem in two ways: First, by adding synthetic mobile screen data containing just image blocks, to equalize the average percentage of blocks of each class per screenshot in the data. And by adding a weighted cross-entropy class loss, each class assigned weight equal to the inverse of the number of samples per class.

\textbf{\emph{3) Training:}}
    End-to-end quantize-aware training is performed on the CNN network. The model is trained on 272 * 480 resolution to maintain a similar aspect ratio as mobile screenshots. However, it supports variable size input for inference. For region proposals for the network, 9 anchor boxes of varying size are used. The center point of the anchor boxes are separated by 16 pixels, making a total of 510 unique center points. The dimensions of the anchor boxes are computed by running k-means algorithm over the annotations of training data. This step is critical as the closer the dimensions of anchor boxes is to the actual data, the better and quicker is the model convergence. For training, we use Adam optimizer with the initial learning rate kept at 0.04 with decay factor of 0.5, decaying after every 10 epochs. Training is halted with the help of early stopping mechanism to avoid overfitting.
    
\begin{figure*}[htbp]
\includegraphics[width=1\textwidth]{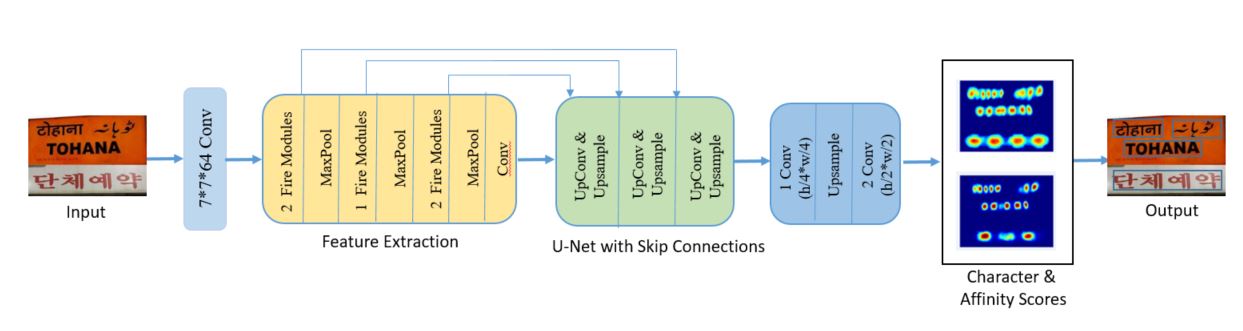}
\label{fig:label}
\centering
\caption{Architecture Diagram for Deep Learning based Text Detection}
\centering
\end{figure*}    
    
\textbf{\emph{4) Weighted NMS:}}
Non-Maximum Suppression (NMS) is an indispensable component in any object detection task. It is used to filter out overlapping proposals and to merge all proposals that belong to the same object. The aim of the NMS module is to suppress the region of detections that do not belong to the object and at the same time not miss or cut any region of proposal that belongs to the object. We apply a novel weighted NMS algorithm to the predicted region propositions, to avoid the disadvantages of the existing greedy-NMS and soft-NMS [17] methods. We safely assume that all the object proposals that overlapped each other with even a very low IOU (set to 0.2 in our case) will belong to the same object. The representative bounding box coordinates ($c_{final}$) for each set of the overlapping proposal is calculated by taking the weighted mean of width, height and center coordinates of all the elements in the set, As depicted by the formula:
\[c_{final}=\frac{(\sum_{(i=1)}^Np_{i}.c_i )}{(\sum_{(i=1)}^Np_i )}\]
Where, N denotes the total number of overlapping boxes with IOU $>$ 0.2, $c_i$ is the geometrical property of the ith box (width, height, center) and $p_i$ is the confidence score of the $i_{th}$ box. 

Detections in the center of the object generally have a higher confidence score in comparison to proposals at the sides. This helps in giving more preference to central proposals at the same time not completely ignoring other proposals. So, the idea is that no one proposal decides or affects the actual boundaries of the object. Empirically, this proved to give a better mAP value for the predicted blocks, than using either greedy NMS or soft NMS. Because in both the conventional methods, if a proposal has IOU less than the NMS threshold, it will also be directly considered as a detection. This can sometimes lead to skewed predictions.

\textbf{\emph{5) On Device Optimizations:}} To support real-time on-device detection of FHD mobile screenshots we perform a number of optimizations. The input is resized to a low resolution of 272 * 480. Quantize-aware training is applied to convert the weights to UINT8 and preserve the same accuracy as the float model.  Post training weight pruning is done to remove redundant weights. Apart from this multi-threading inference and memory reuse is performed, to further boost inference speed. All these optimizations helped us complete the detection and post processing task in less than 50 ms on the Samsung Flagship Galaxy S10 device.

\subsection{Text Block Identification}
Text Extraction refers to the task of identifying and segmenting the regions of text areas in a given input. Extracting text regions can be challenging due to the varied fonts, styles, sizes, shapes, colors, alignment of text. The problem becomes more tedious with multi-colored/ textured background due to the blend of background and foreground text colors. The correct identification of text regions become an important aspect to completing Layout Analysis. Text regions can be further categorized into 3 types:

i)	Independent Text  No dependency on other elements like images or tables

ii)	Overlay Text  Text areas around Images (Caption, etc.)

iii)	Scene Text  text occurring naturally as a part of scene. 

As a known fact, object detection models have a slight deviation in bounding boxes. Thus, the text boxes extracted from the deep learning model can’t be used directly because a slight deviation in detection can cause a cut in the words. Hence, text boxes are identified by one of the two ways mentioned below, depending on the device capabilities and use case requirement. 
Note, that the deviation in image blocks is handled by post-processing and block correction techniques that are discussed later. 

\subsubsection{Image Processing Approach}
In this segment, we focus only on extracting type independent text. Thus, we revert to the image processing techniques to extract text. We follow an approach based on the work of [5],[6],[7] which proposes techniques based on edge detection, thresholding and connected components to identify text regions. Since we have accurately identified the image regions already, we mask them out to save processing and improve the accuracy of text extraction. For the given input, we apply binarization using technique from [8], followed by canny edge detection and morphological operators (dilation/erosion) to emphasize the identified edges. Instead of using Connected Components, we find Contours around the edges based on the algorithm by [9] implemented in image processing library OpenCV. Character to Word to Line level granularity of extracted text regions can be controlled using the morph op Dilation after binarization. As a final step, we prune out the blocks that are of irregular aspect ratio and small size compared to the input. The Steps followed can be described in the following Pseudo Code:
\begin{enumerate}
\item	Mask out the region identified as Image via the neural network from the screenshot to avoid ambiguity caused due to the image region.
\item	After masking as mentioned in point 1, the RGB image is converted to a Binarized Image using method from [8].
\item	Canny Edge Detector is applied, to emphasize the edges on the binarized image.
\item	Contours are identified on the image using OpenCV built-in function, which is converted into a Rectangle Boundary Box.
\item	Intersecting Blocks and Horizontally or Vertically Near-Blocks are merged. Blocks with 
irregular aspect ratio, horizontally small block or too close to the left or right side of screenshot are discarded.
\end{enumerate}
 The above steps give us the result with character level granularity. In case if we wanted a line or word of text instead of an individual character, DILATION can be used with appropriate kernel size before finding contours.

\subsubsection{Deep Learning Based Approach}
As in the image processing approach for text identification, we have focused only on extracting independent text but text in mobile screenshots is not limited to the simple app and web-page text, it can also be present in the form of scene-text embedded on top of image blocks. With this deep learning based text identification approach, we can focus on remained both type of text which was considered as image block in image processing approach. As the deep learning based approach is more resource-intensive so it is recommended for high-end devices. For low-end devices, Image processing approach can be used for text block identification. 

  Both, the deep learning based layout analysis and the image processing based text block identification method are able to detect app and web-page text with less and high accuracy respectively but perform poorly to detect multi-oriented or complex text present on gradient backgrounds or images. Extraction of such kind of text, like text present on a person’s ID or document, text on a meme on social media or a quote on a poster can be very useful for the user. Also, text on image blocks contain some important information about the screenshot like an image of a restaurant board or signboard and hence can be extracted and stored as keywords to support fast image search.
  
  As we support \textbf{Hierarchical Layout Analysis}, i.e. grid blocks can contain multiple text, image or icon blocks. To handle above mentioned scenario, further an image block can contain multiple text blocks present on top of image blocks.

  We create a very light-weight CNN model[19] with a U-net architecture to perform Scene text detection on the mobile screenshot and natural images. It supports the detection of multi-oriented, multi-scale scene text as well as incidental text. Taking motivation from Character Region Awareness for Text Detection[10] (CRAFT), we model the Text Localization task as a semantic segmentation task[11], assigning Gaussian character region scores to localize individual characters and Gaussian affinity scores to group characters of each word together.  
  
  We use a series of fire modules, inspired by SqueezeNet[12] architecture to form the feature extraction block. It is followed by a U-Net architecture to obtain multiscale feature information to strengthen the network'’s feature extraction ability. The U-Net consists of 3 UpConv and UpSample blocks, in which skip connections are fed from the ${2}^{nd}$, ${3}^{rd}$ and ${5}^{th}$ fire module. Each UpConv block consists of 1 * 1 convolutions, followed by a 3 * 3 convolutions. After each UpConv block, we have a resize bilinear, with upscaling factor of 2. The architecture for DL-based text detection is shown in Figure 3.

  The output node if of size h/2 * w/2 * 2, where the 2 channels predict the pixel level character and affinity scores. The scores are fed to a post-processing module that gives word level bounding boxes by extracting connected components from the model score.

\textbf{\emph{Data Annotation:}} By encoding the probabilities of character centers and space centers between adjacent characters, a pixel-level character and affinity score is generated using the method followed by CRAFT. Each of these score maps are modeled as probabilistic Gaussian distributions. In addition to MLT 2019 dataset [13], a synthetic dataset, containing random combinations of myriad scripts in a single image is generated using the process mentioned in SynthText [14] to further bolster training. Also, synthetic screenshot text similar to app and website text is created. A total of 25k images of size 512 x 512 are used for training and 2.5k images are kept for validation and testing. The images are fed to model after performing normalization.

\textbf{\emph{Optimization Function:}} Applying an optimization function directly on pixel-wise deviation in predicted and actual scores may not be suitable because the character and affinity scores at every pixel are highly inter-related with its neighbouring pixels. To tackle this dependence in nearby pixels, we define a novel: logcosh-Pool ($L_{tl}$) cost function given by:

\[L_{tl}(y,y^p)=\sum_{(i=1)}^nlog⁡(cosh⁡(y_{avgpool(s=2)}^p-y_{avgpool(s=2)}))\]
The pooling allows to avoid and cancel out the possible accumulation of small pixel-level deviations in neighboring scores which are irrelevant to us, as the smallest objective is to predict correct character boundaries instead of predicting exact pixel-level scores. The function also gives a smoothening effect to the final output and a boost in the accuracy, giving lesser cuts in word boundaries. 
\begin{figure*}[h!]
\includegraphics[width=1\textwidth]{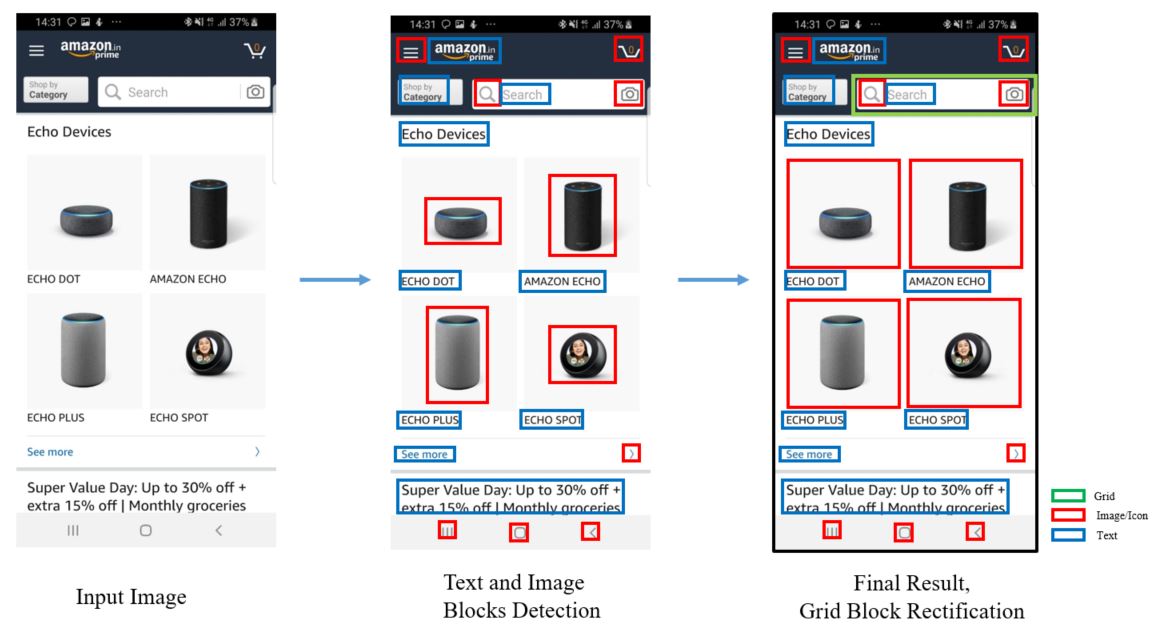}
\label{fig:gridR}
\centering
\caption{Results showing Image Blocks Rectification with Grid Blocks}
\centering
\end{figure*}

\textbf{\emph{On Device Optimizations:}} To speed up Mobile inference and gain significant performance improvement we deploy Quantization Aware Training and convert our model weights and activations to UINT8. We apply knowledge distillation [15] to learn soft probabilities from intermediate layers of the teacher network. Our student network of size 2.4 MB had 65\% fewer parameters than the teacher network of size 7.5 MB. Just training the lower size network with ground truth, we observed around 3.2\% drop in accuracy, whereas when trained with soft probabilities from the teacher network and fine-tuned on ground truth, the drop in accuracy was less than 0.5\%. Further, layer-wise iterative channel pruning [16] and fine-tuning is performed to remove redundant channels and form a highly compressed quantized model of size 800 KB, with 3.3 M parameters with no impact on accuracy. Our scene text localization module runs in 65ms on FHD screenshots on Samsung flagship Galaxy S10 device.

\subsection{Grid Identification}
It is often observed in the layout, that different segments will be separated by a line or any prominent block might be having a boundary. Such boundaries can be considered as a combination of horizontal and vertical edges. To rectify the deviation of observed blocks from neural networks (NN) and image processing (IP) techniques, we devise another algorithm to identify such prominent blocks and call them Grid Blocks. Given the input, we perform a combination of edge detection and morphological operators, after binarization, to obtain horizontal and vertical edges. Canny edge detection fails to identify low intensity edges and results in broken edges. Therefore, we have used the Laplacian method, which use second-order derivative to detect edges. To make sense of extracted edges, we find contours using connected components around them, forming a closed bounding box structure called the Grid. The Pseudo steps can be seen as follows:
\begin{enumerate}

\item	Image is converted to Grayscale, followed by the weighted sum of (a) median blurred gray image, (b) CLAHE based contrast enhanced gray image and (c) Laplacian of gray image. The weights are hyperparameter found using trial-and-error.
\item	The weighted image from the above step is thresholded to get a binarized image.
\item	Bounding Boxes after finding contours are identified.
\item	Boxes with very small areas relative to the image and irregular aspect ratio are removed.
\item	Intersecting blocks are combined using a union operator to form a single block.
\item	If a Grid Block is more than 90\% of image area then remove the Grid Block.
\end{enumerate}
The remaining blocks are labelled as Grid Blocks. These newly formed bounding box or grid blocks are further utilized in two ways:
\subsubsection{Block Rectification}
If the mapping between the grid blocks and previously identified blocks in one-to-one, i.e., the area covered by the grid block is shared by only one previously (NN and IP based) formed block, then we use this block to rectify the former block in its spatial deviation. It is observed that whenever a Grid block is identified, it is accurate due to the accurate estimation of edges and thus being able to rectify the spatial bounding box deviation. It has been illustrated in Figure 4.
\begin{figure*}[h]
\includegraphics[width=1\textwidth]{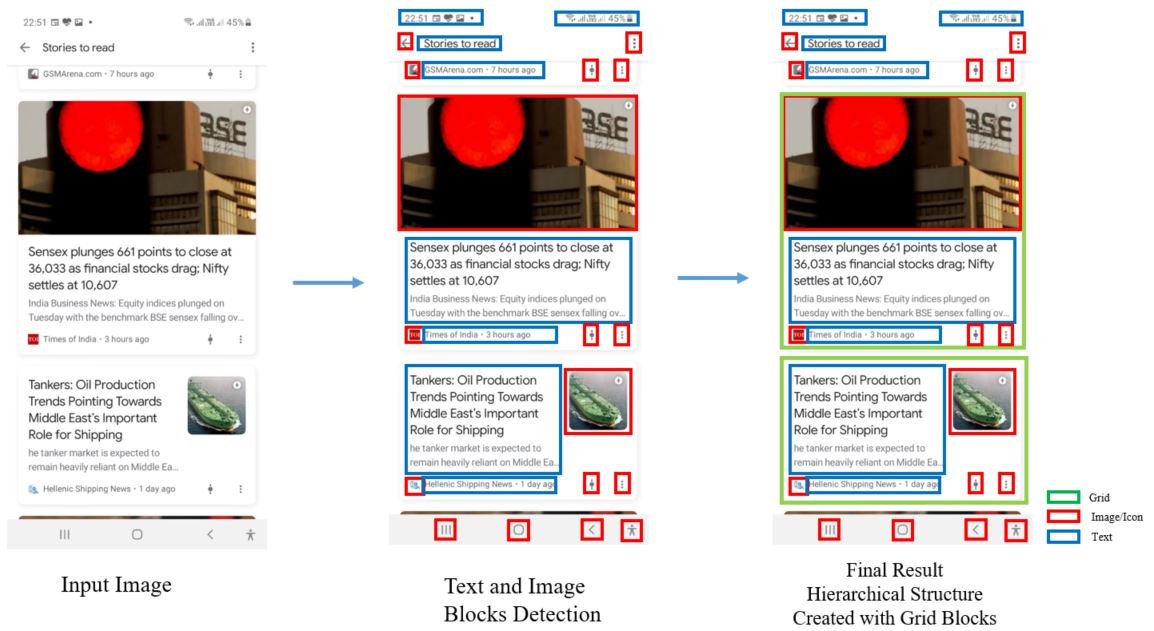}
\label{fig:gridH}
\centering
\caption{Results showing Hierarchical Structure Created with Grid Blocks}
\centering
\end{figure*}
\subsubsection{Block Hiearhirical Structure}
If the mapping between the grid blocks and previously identified blocks are not one-to-one, i.e., the area covered by the grid block is shared by more than one previously (NN and IP based) formed block, then we identify the newly formed block as the segment of the layout and blocks that it contains becomes the sub-segmentation of the identified segment. As shown in Figure 5, providing us with a hierarchical view of the layout.

\section{Results}
As there is no standard or readily-available dataset for testing of our solution. We evaluate our architecture on a custom test dataset of collected 750 images of different varieties which contains screenshots, scene text and wallpaper type of images. This dataset is hand-annotated by 3 trained annotators. The screenshot test images from the dataset cover a wide variety of screenshots taken across various web and app domains like Shopping, Social Media, Payments, Food and Native apps. The Sample output of input screenshots is shown in Figure 4 and 5. The predicted bounding blocks of image and icon class are drawn in red while the blocks of text class are drawn in blue. On the top of those blocks, grid block is shown in green.

\begin{figure}[h!]
\includegraphics[width=1\linewidth]{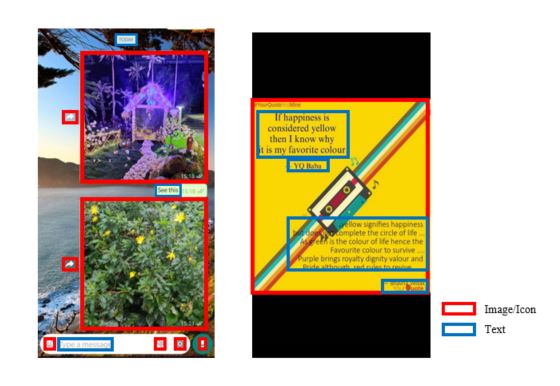}
\label{fig:resS}
\centering
\caption{ Success results on some of the complex scenarios: In Left Image, detection of image on top of image. In right one, detection of poster images which has comparatively less image characteristics. These cases require foreground-background and semantic understanding of the image respectively}
\centering
\end{figure}
  
  To measure the accuracy, we have calculated IoU (Intersection Over union) between Ground Truth and predicted result to determine the unidentified regions or deviations in a given screenshot. IoU measures the overlap between the two boundaries.We define an IoU threshold $\geq$ 0.75 (which is consider as strict metrics to analyze accuracy of layout analysis solution) in classifying whether the prediction is a true positive or a false positive. 
  
  The proposed optimized on device capable pipeline achieves an average precision value of 0.95. While running solution on a Samsung Galaxy device S10 having processor EXYNOS 9820, pipeline is observed to run with an average of 200 ms for a screenshot image of size 1080p * 2240p. 60 MB RAM is enough to run the full pipeline on mobile devices. The full pipeline requires only 3 MB of ROM.
  
  The prediction accuracies for each class is evaluated separately and shown in Table 1. It can be seen that suggested approach achieves the best results for all classes i.e. icon, image and text.

\begin{table}[htpb]
\centering
\caption{Experimental result of ScreenSeg on test dataset}
\begin{tabular}[t]{ccccc}
\toprule
Class & Precision& Recall & HMean\\
\toprule
Image & 0.93 & 0.96 & 0.94 \\
Text  & 0.97 & 0.93 & 0.95 \\
Icon  & 0.96 & 0.92 & 0.94 \\
\bottomrule
\textbf{Overall} & \textbf{0.95}& \textbf{0.93} & \textbf{0.94}\\
\bottomrule
\end{tabular}
\end{table}

\begin{table}[htpb]
\centering
\caption{Experimental Result on different categories of images}
\begin{tabular}[t]{ccccc}
\toprule
Class & Precision& Recall & HMean\\
\toprule
Screenshot Images & 0.94 & 0.93 & 0.94 \\ Scene Images & 0.98 & 0.94 & 0.96 \\
Wallpaper Images & 0.92 & 0.90 & 0.91 \\
\end{tabular}
\end{table}

  \begin{figure}[h!]
\includegraphics[width=1\linewidth]{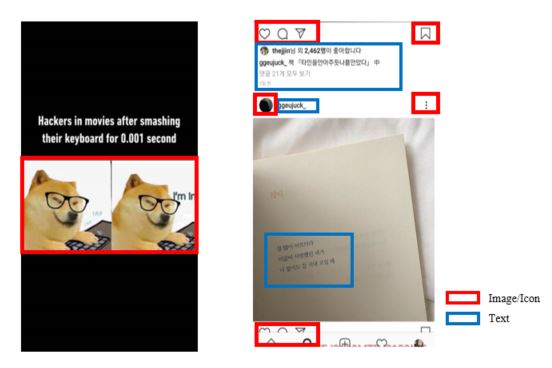}
\label{fig:resF}
\centering
\caption{ Failure scenarios: In Left Image, Its not detecting semantic block with Text and Image combined together, as no visible boundary is combining image with text. In Right one, image block is missed, as there is no foreground background color variation and very less image characterstics.}
\centering
\end{figure}
 
\section{Conclusion and Future Work}
We introduced a novel and optimized on-device pipeline for layout analysis of screenshot. The solution pipeline is already commercialized in various applications of Samsung Device. This type of solution is not yet available in competitor devices. Google lens provides selection of text regions and supports similar image search from a screenshot, but the solution is server based and requires an internet connection. Also, to the best of our knowledge, it doesn't support multi-level hierarchy and grid block extraction.
 
To summarize, this paper describes the procedures followed for layout analysis of screenshots on-device by combining Deep Neural network and Standard Vision Analysis techniques. The whole pipeline runs in 200 ms and has a precision value of 0.95 and a recall value of 0.93. The results are promising and have a wide-range of use cases including Smart Crop, MyFilter etc. This pipeline is used for layout analysis, semantic block identification and symmetricity identification and can be extended to InPlace Text Extraction or improving the accuracy of OCR Engine with detected text blocks.

  Layout Analysis of mobile screenshots can be expanded to support more subclasses, i.e. Image class can further be classified as Posters, Promotions, Natural Images etc. and can help highlight only important information. We also plan to extend our approach to detect and efficiently extract information from Tables. Intelligent extraction and ranking based on usefulness of the blocks is also part of our future work. Currently, we rank blocks based on their position and geometric properties, it can be extended to consider semantics and block content.

\end{document}